\pdfoutput=1
\documentclass[11pt]{article}
\usepackage{emnlp2021}
\usepackage{times}
\usepackage{latexsym}
\usepackage{amsmath}

\usepackage{xcolor, colortbl}
\usepackage[ruled,vlined,linesnumbered]{algorithm2e}
\usepackage[noabbrev,capitalize,nameinlink]{cleveref}
\usepackage{multirow, graphicx}
\usepackage{booktabs}
\usepackage{textcomp}
\usepackage{float}

\definecolor{Gray}{gray}{0.85}

\usepackage[T1]{fontenc}
\usepackage[utf8]{inputenc}
\usepackage{microtype}

\title{Cartography Active Learning}

\author{
  Mike Zhang \and Barbara Plank \\
  Department of Computer Science \\
  IT University of Copenhagen \\
  \texttt{\{mikz, bapl\}@itu.dk}}

\begin{document}
\maketitle
\begin{abstract}
We propose Cartography Active Learning (CAL), a novel Active Learning (AL) algorithm that exploits  the behavior of the model on individual instances \textit{during training} as a proxy to find the most informative instances for labeling. CAL is inspired by data maps, which were recently proposed to derive insights into dataset quality~\citep{swayamdipta-etal-2020-dataset}.
We compare our method on popular text classification tasks to commonly used AL strategies, which instead rely on post-training behavior. We demonstrate that CAL is competitive to other common AL methods, showing that training dynamics derived from small seed data can be successfully used for AL. We provide insights into our new AL method by analyzing batch-level statistics utilizing the data maps. Our results further show that CAL results in a more data-efficient learning strategy, achieving comparable or better results with considerably less training data.
\end{abstract}
\section{Introduction}
Active Learning (AL) is a widely-used method to tackle the time-consuming and expensive collection and manual labeling of data. 
In recent years,  many AL strategies were proposed. The simplest and most widely used is \textit{uncertainty sampling}~\cite{lewis1994sequential, lewis1994heterogeneous}, where the learner queries instances that it is most uncertain about. Uncertainty sampling is myopic:\ it only measures the information content of a single data instance.
Alternative AL algorithms instead focus on selecting a \textit{diverse} batch~\citep{geifman2017deep, sener2017active, gissin2019discriminative, zhdanov2019diverse} or to estimate the \textit{uncertainty} distribution of the learner~\citep{houlsby2011bayesian, gal2016dropout}. However, these methods are usually limited in their notion of informativeness, which is tied to post-training model uncertainty and batch diversity.  

\begin{figure*}[htbp]
    \centering
    \includegraphics[width=.45\linewidth]{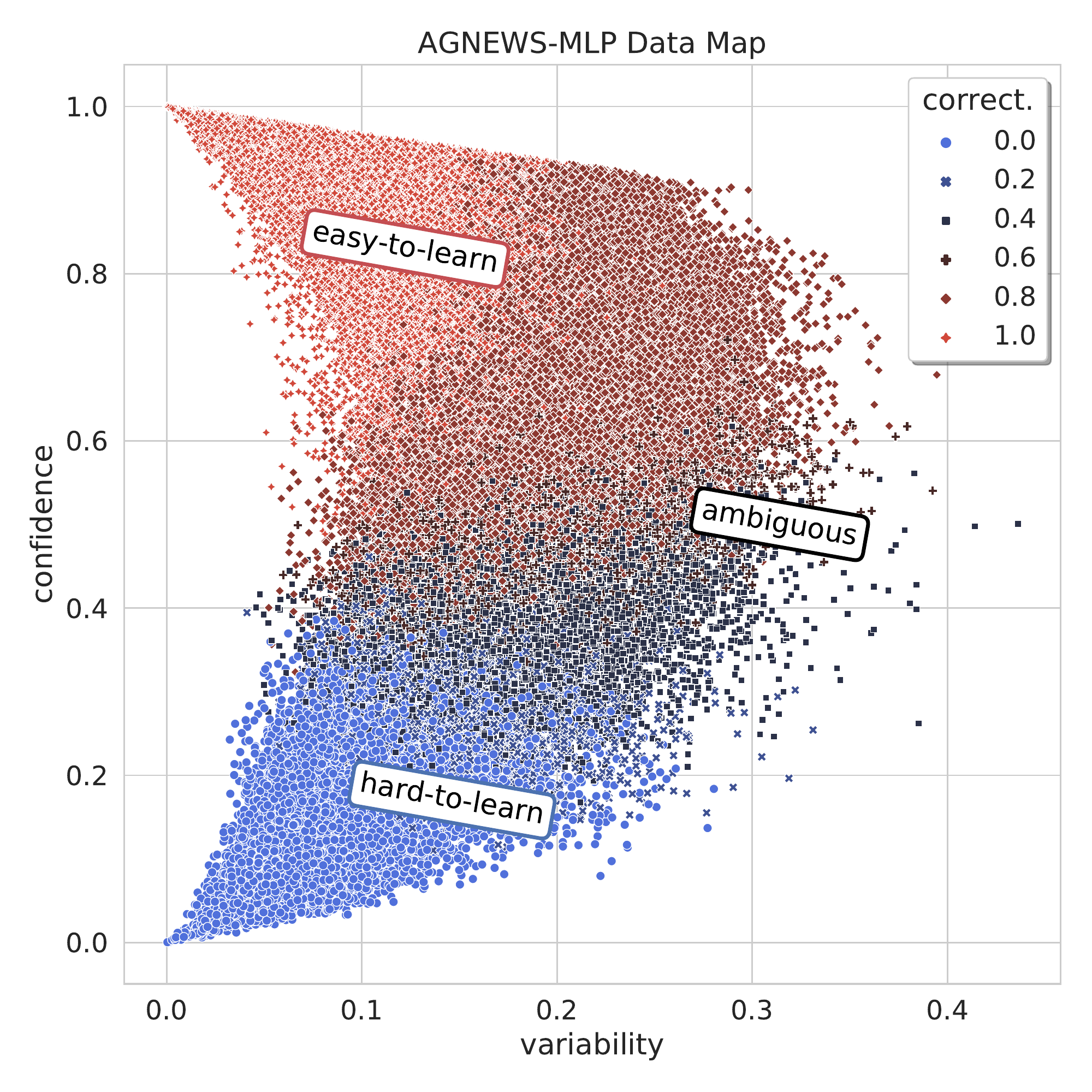}
    \hspace{1em}
    \includegraphics[width=.45\linewidth]{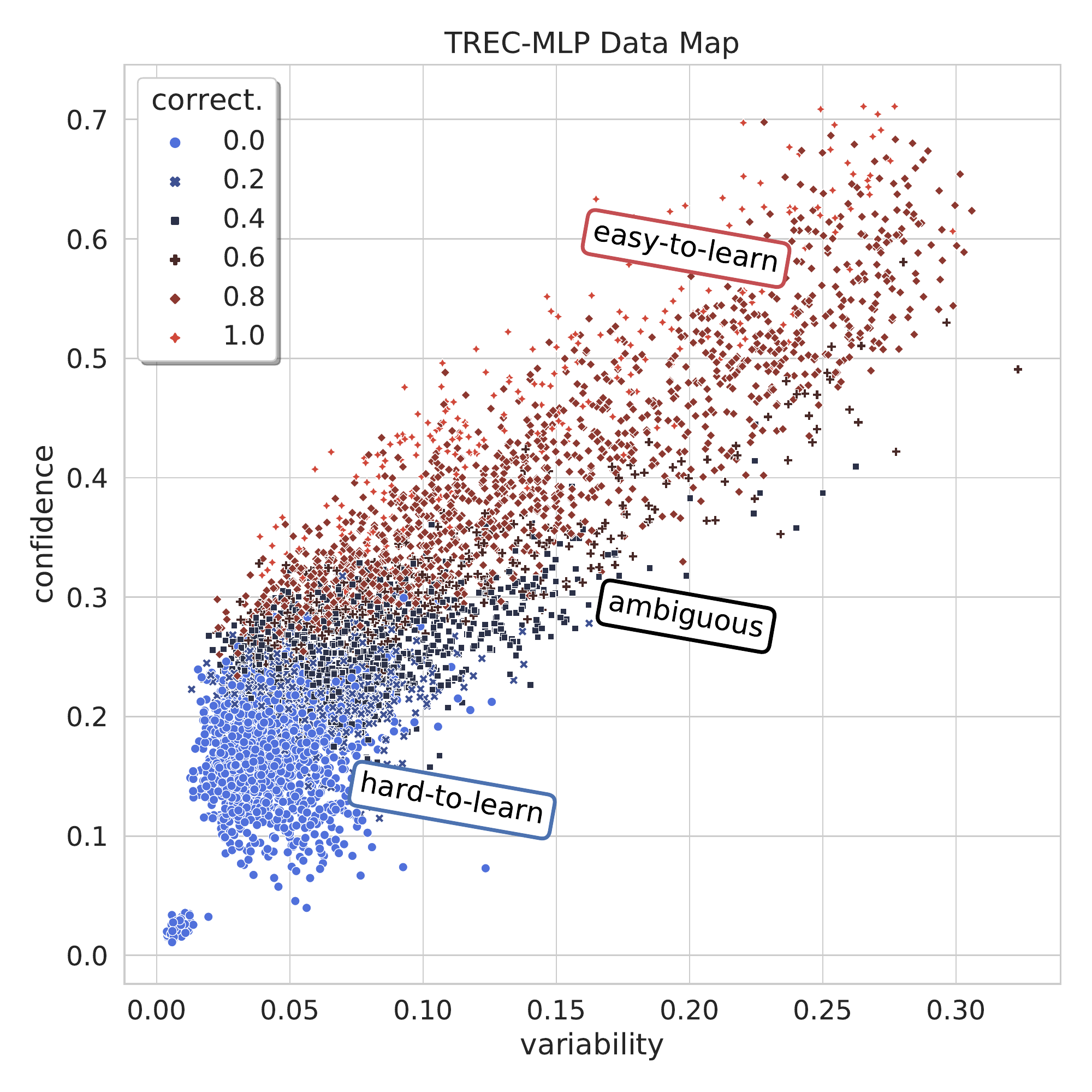}
     \caption{\textbf{Full Data Maps for AGNews \& TREC.} AGNews (120,000 instances) on the left, and  TREC  (5,452 instances) on the right, both w.r.t.\ an MLP training for ten epochs.  The x-axis shows \textbf{variability} and the y-axis the \textbf{confidence}. The colors and shapes indicate the \textbf{correctness}.}
    \label{fig:trec-agnews}
\end{figure*}

Recently, \citet{swayamdipta-etal-2020-dataset} introduced \textit{data maps}, 
to visualize the behaviour of the model on individual instances during training (\textit{training dynamics}). The plotted data maps (\cref{fig:trec-agnews}) reveal distinct regions in a dataset:\ groups of \textit{ambiguous} instances useful for high performance and linked to high informativeness, \textit{easy-to-learn} instances which aid optimization, and \textit{hard-to-learn} instances which frequently correspond to mislabeled or erroneous instances.

We propose \textit{Cartography Active Learning} (CAL), which automatically selects the the most informative instances that contribute optimally to model learning. To do so, we leverage a largely ignored source of information:\  insights derived \textit{during training}, i.e., training dynamics derived from \textit{limited} data maps (see~\cref{sec:method}) to choose informative instances at the boundary of ambiguous and hard-to-learn instances. We hypothesize that this region is where the model will \textit{learn the most} from.
Data maps provide the additional benefit that we can use them to measure informativeness of a batch with straightforward metrics and visualize dataset properties. These distinct regions in the data maps have their own respective statistics. Therefore, as a second research question we investigate whether data map statistics help to assess why some AL algorithms work better than others.

\paragraph{Contributions} 
In this paper, our contributions are twofold. (1) We present Cartography Active Learning, a novel AL algorithm that exploits data maps for AL. 
We compare our results against other competitive and widely used AL algorithms and outperform them in early AL iterations. (2) Additionally, we leverage the data maps 
to inspect what instances AL methods select. We show that our approach optimally selects informative instances avoiding only \textit{hard-to-learn} and \textit{easy-to-learn} cases, which leads to better AL and comparable or better results than full dataset training. 

\section{Related Work}\label{background}
 AL has seen many usage scenarios in the Natural Language Processing (NLP) field~\citep{shen2017deep,lowell-etal-2019-practical, ein-dor-etal-2020-active, margatina2021active}. The perspective of AL is that if a model is allowed to select the data from which it will \textit{learn the most}, it will achieve comparable (or better) performance with less training instances~\cite{siddhant2018deep}, and at the same time addressing the costly labeling process with a human annotator.

A popular scenario is pool-based active learning~\citep{lewis1994sequential,settles2009active, settles2012active}, which assumes a small set of labeled data $\,\mathcal{L}$ and a large pool of unlabeled data  $\,\mathcal{U}$.
Most AL algorithms start similarly:\ a model is fit to  $\,\mathcal{L}$ to get access to $P_\theta(y\mid \boldsymbol{x})$, then apply a query strategy to get the best scored instance from  $\,\mathcal{U}$, label this instance and add it to  $\,\mathcal{L}$ in an iterative process.

\paragraph{Common Strategies} A commonly used query strategy is \textit{uncertainty sampling}~\cite{lewis1994sequential, lewis1994heterogeneous}. In this approach, the learner queries the instances which it is least certain about.
There are two popular approaches. (1) Uncertainty sampling based on entropy~\citep{shannon1948mathematical, dagan1995committee}, it uses the entropy of the label distribution as a measure for the uncertainty of the model on an instance. (2) Uncertainty sampling based on which best labeling is the least confident~\citep{culotta2005reducing}.

\paragraph{Batch-mode Active Learning} It is inefficient and time-consuming to obtain sampled queries one by one for annotation in the context of Deep Neural Networks (DNNs).
In a real-world setting, consider having multiple annotators available. One can exploit this setting and label the instances in batches and parallel.
Batch-mode AL allows the learner to query instances in groups. To assemble the optimal batch, one can greedily pick the top-$k$ examples according to an instance-level acquisition function suitable for DNNs. 
There are many works on ways for making neural network posteriors accurately represent the confidence on a given example. One popular example is stochastic regularisation techniques such as dropout during inference time, known as the Monte Carlo Dropout technique~\citep{houlsby2011bayesian}.~\citet{gal2016dropout} refer to this as Bayesian Active Learning by Disagreement (BALD). This allows us to consider the model as a Bayesian neural network and calculate approximations of uncertainty estimates by analyzing its multiple predictions. However, if the information of these top-$k$ examples is similar, this will result in the model not generalizing well over the dataset. Therefore, alternative approaches take the \textit{diversity} of a batch into account.

\paragraph{Batch-aware Query Strategies}
Instead of greedily choosing the examples that maximize some score, one can instead try to find a batch that is as diverse as possible. 
One recently proposed effective strategy is Discriminative Active Learning~\citep[DAL;\ ][]{gissin2019discriminative}. This approach aims to select instances from $\,\mathcal{U}$ that make $\,\mathcal{L}$ representative of $\,\mathcal{U}$. In other words, the idea is to train a separate model to classify between $\,\mathcal{L}$ and $\,\mathcal{U}$. Then, to use that model to choose the instances which are most confidently classified as being from $\,\mathcal{U}$. 
If $\,\mathcal{U}$ and $\,\mathcal{L}$ become indistinguishable, the learner has successfully closed the data gap between $\,\mathcal{U}$ and $\,\mathcal{L}$. DAL was proposed for computer vision and was recently successfully used in NLP~\cite{ein-dor-etal-2020-active}. Alternative diversity AL strategies exist, such as core-set, which often rely on heuristics~\citep{sener2017active, geifman2017deep}.\footnote{Contemporary to our work and both appearing at EMNLP 2021,~\citet{margatina2021active} proposed CAL (Contrastive Active Learning), another AL method with the same acronym to our approach. The main difference to Cartography AL is using data points that are similar in the model feature space, while optimizing for maximally disagreeing predictive likelihoods.}

\section{Cartography Active Learning}\label{cal}
The key idea of CAL is to use model-independent measures, from fitting the model on the seed data $\,\mathcal{L}$, by using data maps~\citep{swayamdipta-etal-2020-dataset} for AL. Data maps help identify characteristics of instances within the broader trends of a dataset by leveraging their training dynamics (i.e., the behavior of a model \textit{during training}, such as mean and standard deviation of confidence and correctness with respect to the gold label). These model-dependent measures reveal distinct regions in a data map, by and large, reflecting instance properties (see~\cref{fig:trec-agnews} and details below on \textit{easy-to-learn}, \textit{ambiguous}, and \textit{hard-to-learn} instances). Training dynamics encapsulate information of data quality that has been largely ignored in AL:\ the sweet spot of instances at the boundary of \textit{hard-to-learn} and \textit{ambiguous} instances, which are quick to label while providing informative samples, as shown in full data training~\citep{swayamdipta-etal-2020-dataset}.

In the next part, we introduce training characteristics, first showing the resulting data maps on the full data. Then we introduce CAL, which proposes to learn a data map from the seed labeled data $\,\mathcal{L}$ and identifying regions of instances with a binary classifier, inspired by DAL~\citep{gissin2019discriminative}. To identify these regions, we require:\ (1) a data map can be learned from limited data, and (2) a classifier to identify informative instances. The full algorithm, illustrated in~\cref{algo:cal}, is described later. 

\begin{figure*}[ht]
    \centering
    \includegraphics[width=.45\linewidth]{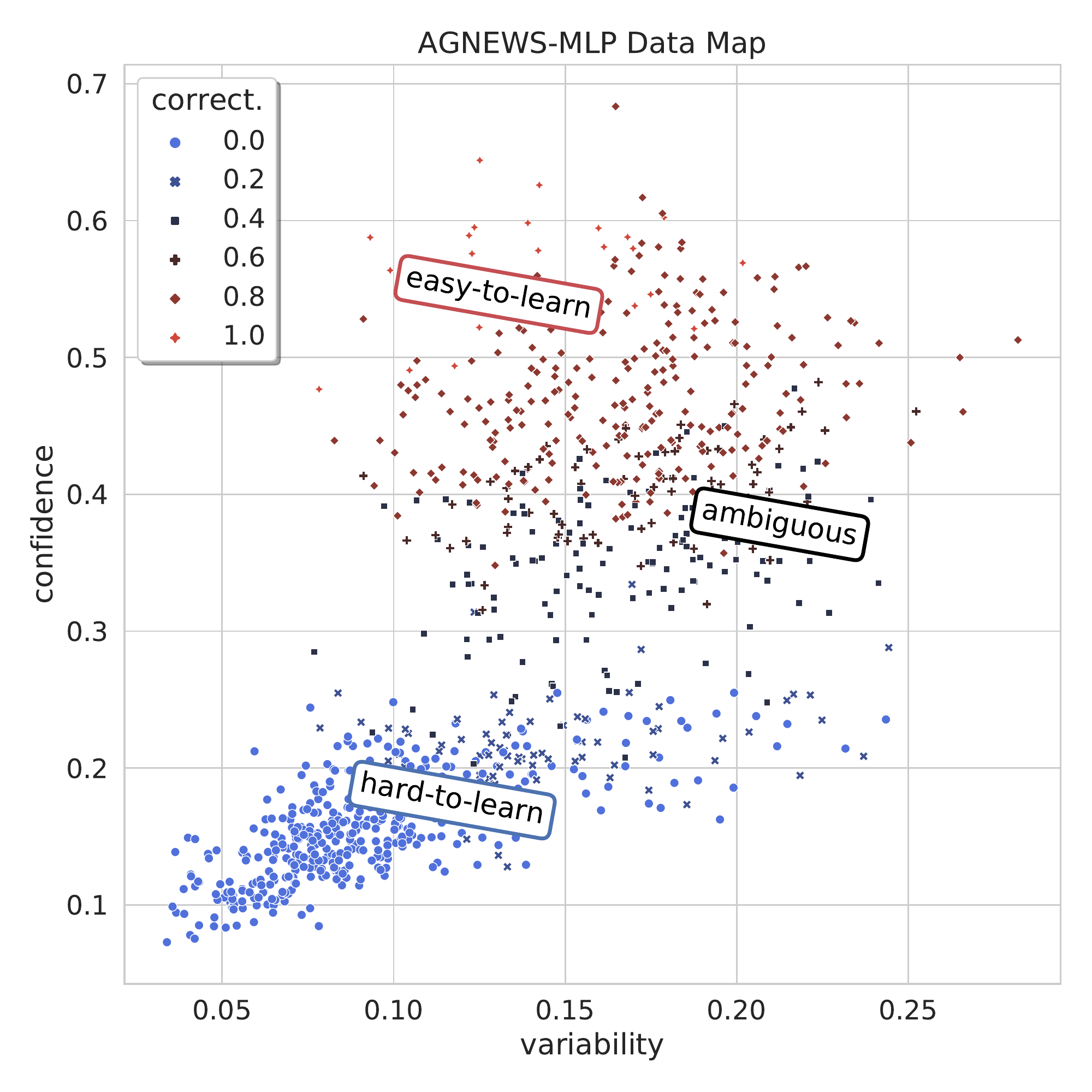}
    \hspace{1em}
    \includegraphics[width=.45\linewidth]{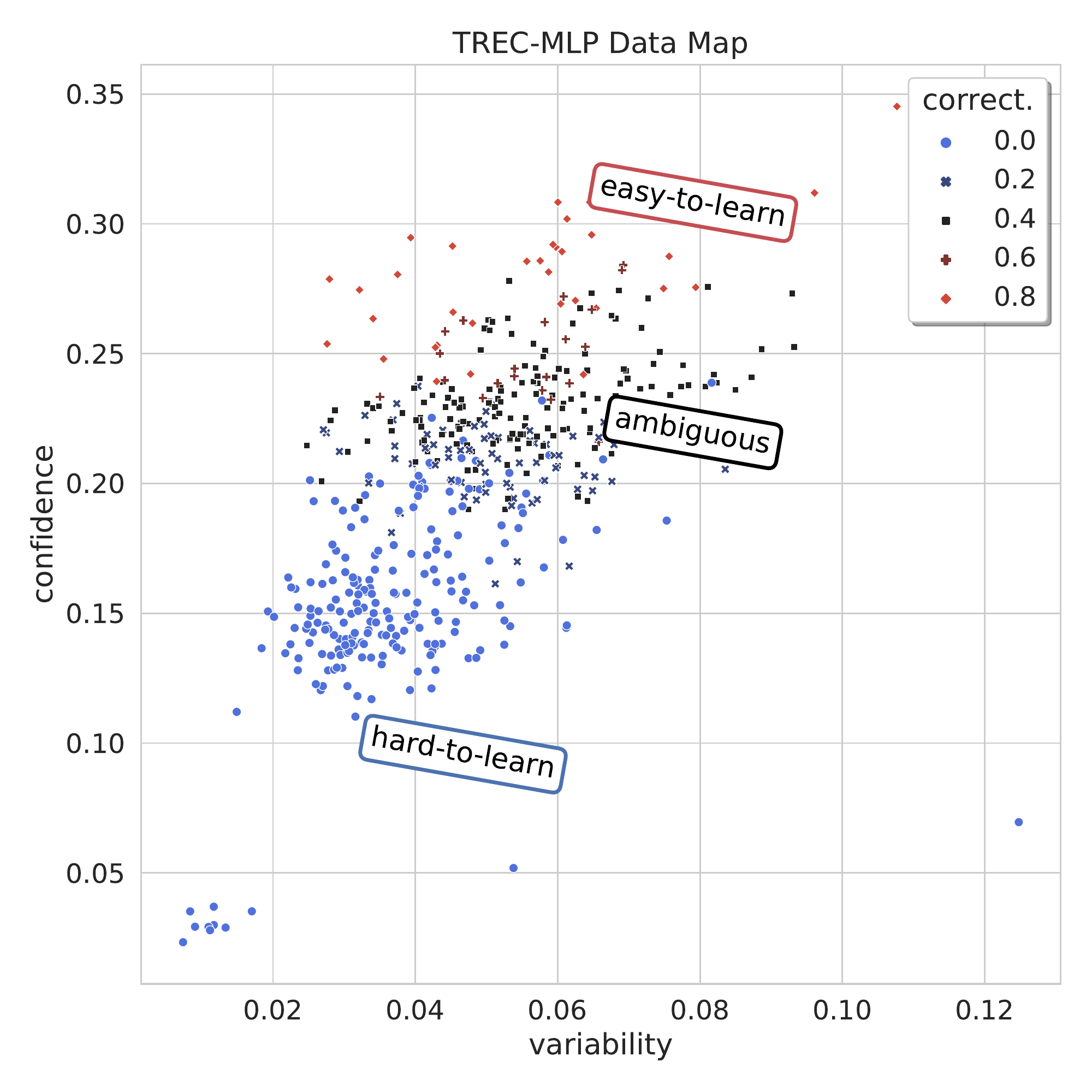}
    \caption{\textbf{Data Maps with Limited Seed Data.} Data map for the AGNews seed set (1,000 instances), and TREC seed set (500 instances). Both data maps are based on an MLP trained for ten epochs.}
    \label{fig:trec-agnews-seed}
\end{figure*}

\paragraph{Mapping the Data}
Formally, the training dynamics of instance $i$ are defined as the statistics calculated over $E$ epochs. These statistics are then used as the coordinates in the plot. The following statistics are calculated, \textbf{confidence}, \textbf{variability}, and \textbf{correctness}, following the notation of~\citet{swayamdipta-etal-2020-dataset}:

\begin{equation}{\label{eq:conf}}
    \hat{\mu}_{i}=\frac{1}{E} \sum_{e=1}^{E} p_{\boldsymbol{\theta}^{(e)}}(y_{i}^{*} \mid \boldsymbol{x}_{i})
\end{equation}
\textbf{Confidence}\footnote{Similar to~\citet{swayamdipta-etal-2020-dataset}, we note that the term \textit{confidence} here is the output probability of the model over the gold label as opposed to the certainty of the predicted label as commonly used in AL literature.}~(\autoref{eq:conf}) is the mean model probability of the gold label $(y_{i}^{*})$ across epochs. Where $p_{\boldsymbol{\theta}^{(e)}}$ is the model's probability with parameters $\boldsymbol{\theta}^{(e)}$ at the end of the $e$\textsuperscript{th} epoch.

\begin{equation}{\label{eq:var}}
    \hat{\sigma}_{i}=\sqrt{\frac{\sum_{e=1}^{E}\big(p_{\boldsymbol{\theta}^{(e)}}(y_{i}^{*} \mid \boldsymbol{x}_{i})-\hat{\mu}_{i}\big)^2}{E}}
\end{equation}
Then, \textbf{variability}~(\autoref{eq:var}) is calculated as the standard deviation of $p_{\boldsymbol{\theta}^{(e)}}(y_{i}^{*} \mid \boldsymbol{x}_{i})$, the spread across epochs $E$.

\begin{equation}{\label{eq:corr}}
    \hat{\phi_i} = \frac{1}{E}\sum_{e=1}^{E}\boldsymbol{1}(\hat{y_i} = y_{i}^{*} \mid \boldsymbol{x}_{i})
\end{equation}

Last, \textbf{correctness}~(\autoref{eq:corr}) is denoted as the fraction of times the model correctly labels instance $\boldsymbol{x}_{i}$ across epochs $E$.

Given the aforementioned training dynamics and the obtained statistics per instance, we plot the data maps for both AGNews~\citep{Zhang2015CharacterlevelCN} and TREC~\citep{li-roth-2002-learning}, using all training data (\cref{fig:trec-agnews}). The data map is based on a Multi-layer Perceptron (MLP).
As shown by~\citet{swayamdipta-etal-2020-dataset}, data maps identify three distinct regions:\ \textit{easy-to-learn}, \textit{ambiguous}, and \textit{hard-to-learn}. The \textit{easy-to-learn} instances are consistently predicted correctly with high confidence, these instances can be found in the upper region of the plot. The \textit{ambiguous} samples have high variability and the model is inconsistent in predicting these correctly (middle region). The instances that are (almost) never predicted correctly, and have low confidence and variability, are referred to as \textit{hard-to-learn} cases. This confirms findings by
\citet{swayamdipta-etal-2020-dataset} where they show that training on the samples of these distinct regions, and in particular the \textit{ambiguous} instances, promote optimal performance. 
While uncertainty-based AL mostly focus on hard-cases, CAL instead focuses on ambiguous and possibly easier instances.

\paragraph{Data Maps from Seed Data}
Given the distinct regions for data selection in the full data map in~\cref{fig:trec-agnews}, we first investigate whether regions are still identifiable if we have little amounts of training data, as this is a prerequisite for CAL.~\cref{fig:trec-agnews-seed} shows this for 1,000 training samples of AGNews and 500 of TREC. We can see that the data points are more scattered, where the \textit{easy-to-learn} and \textit{ambiguous} samples are mixed. However, it seems the \textit{hard-to-learn} region can still qualitatively be distinguished from the other regions.

\paragraph{CAL Algorithm} The algorithm is detailed in~\cref{algo:cal} and described next. To select presumably informative instances, we train a binary classifier on the seed set $\,\mathcal{L}$ and apply it to $\,\mathcal{U}$ to select the instances that are the closest to the decision boundary between \textit{ambiguous} and \textit{hard-to-learn} instances. By visualizing a decision function in~\cref{fig:trec-agnews-seed} that separates the \textit{hard-to-learn} region from the \textit{ambiguous}/\textit{easy-to-learn} region, we select the instances that are the closest to this boundary. In other words, selecting instances with output probability 0.5 with respect to the binary classifier. This does two things, (1) it prevents the binary classifier from selecting only \textit{easy-to-learn} instances (low-variability, high-confidence), and (2) selecting some truly \textit{hard-to-learn} instances (low-variability, low-confidence) which are both not optimal for learning.

Similar to DAL, for the binary classification task, we map our original input space $\,\mathcal{X}$ to the learned representation~$\Psi$ of the last hidden layer of an MLP (\cref{configurations}). These are the features used in our binary classifier ($\theta'$). Formally, as we have three hidden layers, \[\Psi : \mathcal{X} \rightarrow \mathcal{\hat{X}},
\text{where } \Psi = \mathbf{h}_3 = f( W_3 \cdot \mathbf{h}_2 + \mathbf{b}_3).\]
For label space $\mathcal{Y}$, we consider the binary values $\{0,1\}$. This label depends on the \textit{correctness}.
The label $y_{\Psi(\hat{\boldsymbol{x}}_i)}$ for the learned representation of instance ${\hat{\boldsymbol{x}}_i}$ is labeled 1 when the \textit{correctness} using the limited data map at epoch $E$ is above the threshold $t_{cor} > 0.2$. We refer to these as \textbf{high-cor} cases. The samples that are rarely correct ($t_{cor}$ $\leq 0.2$) are labeled as 0 and we refer to these as \textbf{low-cor} cases. To give a better intuition, we refer to~\cref{fig:trec-agnews-seed}, where the regions of \textit{hard-to-learn} and \textit{ambiguous/easy-to-learn} are visually separable with this \textit{correctness} threshold $t_{cor} =$ 0.2. This threshold is empirically chosen by investigating the influence of different \textit{correctness} thresholds on the performance of CAL in~\cref{analysis:threshold} (\cref{tab:threshold}).

\begin{algorithm}[htbp]
\SetAlgoLined
 \textbf{input:} Labeled seed set $\mathcal{L}$, Unlabeled set $\mathcal{U}$, Total budget $K$, Number of queries $n$, Correctness threshold $t_{cor}=0.2$;\\
\For{\text{i = 1, ..., n}}{
 $\Psi(\mathcal{L})$, $\Psi(\mathcal{U}) \leftarrow$ \text{train main classifier} $\theta$ \text{on} $\mathcal{L} \text{, get representations of } \mathcal{L} \text{ and } \mathcal{U}$\;
  $\hat{\mu}$, $\hat{\sigma}$, $\hat{\phi}$ $\leftarrow$ \text{get data map statistics of} $\mathcal{L}$ \text{with} $\theta$\;
  $P_{\theta^\prime}\leftarrow$\text{ train binary classifier $\theta^\prime$ on }$\Psi(\mathcal{L})$
  \text{ with }  $y_{\Psi(\hat{\boldsymbol{x}}_i)} = \begin{cases}
    1,      & \text{if } \hat{\phi_i} > t_{cor}\\
    0,      & \text{else}
\end{cases}$\\
 \For{\text{j=1, ..., $\frac{K}{n}$}}{
  $\hat{\boldsymbol{x}} \leftarrow \underset{\boldsymbol{x}\in \Psi(\mathcal{U})}{\mathrm{argmin}}\, \lvert 0.5 - P_{\theta^\prime}(\hat{y} = 1 \mid \boldsymbol{x}) \rvert$\;
  $\mathcal{L} \leftarrow \mathcal{L} \cup \hat{\boldsymbol{x}}$\;
  $\mathcal{U} \leftarrow \mathcal{U} \backslash \hat{\boldsymbol{x}}$\;
 }
 \text{reset parameters $\theta$ and $\theta^\prime$}\;
  \textbf{return} $\mathcal{L}$, $\mathcal{U}$
 }
 \caption{Cartography Active Learning}
 \label{algo:cal}
\end{algorithm}

\section{Experimental Setup}\label{sec:method}
We focus on pool-based active learning.
Once trained on a seed set $\,\mathcal{L}$, we begin the simulated AL loop by iteratively selecting instances based on the scoring of an acquisition function. We take the top-50 instances, following prior work~\cite{gissin2019discriminative,ein-dor-etal-2020-active}. The selected instances are shown the withheld label and added to the labeled set $\,\mathcal{L}$ and removed from $\,\mathcal{U}$. We evaluate the performance of the trained model on a predefined held-out test set.
We run 30 AL iterations, over five random seeds, and report averages over these runs.

\subsection{Datasets}\label{sec:data}
\begin{table}[htbp]
    \centering
    \resizebox{\linewidth}{!}{
    \begin{tabular}{l|rrrr}
    \toprule
        Dataset & Train     & Test      & Classes   & Seed set size \\ \midrule
        AGNews  & 120,000   & 7,600     & 4         & 1,000\\
        TREC    & 5,452     & 500       & 6         & 500\\\bottomrule
    \end{tabular}}
    \caption{\textbf{Datasets.} Statistics of the two datasets.}
    \label{tab:datastat}
\end{table}
\noindent
In our AL setup, we consider two popular text classification tasks, namely \textbf{AGNews}~\citep{Zhang2015CharacterlevelCN} and \textbf{TREC}~\citep{li-roth-2002-learning}.\ The AGNews task entails classifying news articles into four classes:\ world, sports, business, science/technology. For TREC, the task is to categorize questions into one of six categories based on the subject of the question, such as questions about locations, persons, concepts, et cetera.
Statistics of the data can be found in~\cref{tab:datastat}.
We start with a seed set size of 1,000 for AGNews and 500 for TREC, both are stratified. This means after the AL iterations we will have 2,500 labeled instances for AGNews and 2,000 for TREC. Our motivation here is to keep the AL simulation realistic. We assume enough annotation budget to initially annotate 500--1,000 samples. Then, in every AL iteration annotate an additional 50 samples, which seems manageable for an annotator. Finally, we run 30 AL iterations to give a good overview of the performance of the acquisition functions over the iterations towards convergence.

\subsection{Acquisition Functions}\label{subsec:al}
We consider five acquisition functions.
We opt for a random sampling baseline (\textbf{Rand.}), four existing acquisition functions, and our proposed CAL algorithm. We chose these as they are state-of-the-art and cover a spectrum of acquisition functions (uncertainty, batch-mode and diversity-based). 

\paragraph{Least Confidence~\citep[LC;\ ][]{culotta2005reducing}} It takes 
\[\underset{\boldsymbol{x}\in \mathcal{U}}{\mathrm{argmax}}\, 1 - P_\theta(\hat{y}\mid \boldsymbol{x})\] 
of the predictive (e.g.\ softmax) distribution as the model's uncertainty, and chooses instances with lowest predicted probability.

\paragraph{Max-Entropy \citep[Ent.;\ ][]{dagan1995committee}} Another popular example is entropy based sampling. Instances are selected according to the function
\[\underset{\boldsymbol{x}\in \mathcal{U}}{\mathrm{argmax}}\,- \sum_{y \in\mathcal{Y}} P_\theta(y\mid \boldsymbol{x}) \log_2 \big(P_\theta(y\mid \boldsymbol{x})\big)\]
and again, based on the \textit{a posteriori} probability distribution.

\paragraph{Bayesian Active Learning by Disagreement \citep[BALD;\ ][]{houlsby2011bayesian, gal2016dropout}} This approach entails applying dropout at test time, then estimating uncertainty as the disagreement between outputs realized via multiple passes through the model. We use the Monte Carlo Dropout technique on ten inference cycles, with the max-entropy acquisition function.

\paragraph{Discriminative Active Learning \citep[DAL;\ ][]{gissin2019discriminative}} This approach poses AL as a binary classification task, it uses a separate binary classifier as a proxy to select instances that make $\mathcal{L}$ representative of the entire dataset (i.e., making the labeled set indistinguishable from the unlabeled pool set). The input space for the binary classifier is task-agnostic. One maps the original input space $\mathcal{X}$ to a learned representation $\hat{\mathcal{X}}$ as the input space, with label space $\,\mathcal{Y} = \{l,u\}$ referring to labeled and unlabeled. In the original paper, the learned representation is defined as the logits of the last hidden layer of the main classifier which solves the original task.
Formally, it selects the top-$k$ instances that satisfy
\[\underset{\boldsymbol{x}\in \mathcal{U}}{\mathrm{argmax}}\, \hat{P_\theta}(\hat{y} = u\mid \Psi(\boldsymbol{x}))\]
where $\hat{P_\theta}$ is the trained binary classifier given the learned representations $\Psi$ of instances $\boldsymbol{x}$. 

\subsection{Configurations}\label{configurations}
This work uses two models for the AL setup solving the classification tasks. All the code is open source and available to reproduce our results.\footnote{\url{https://github.com/jjzha/cal}}
\paragraph{Main Classifier}
We use a Multi-layer Perceptron (MLP), with three $d_{\text{emb}}= $ 300 ReLu layers, dropout probability $p = 0.3$, minimize cross-entropy,
AdamW optimizer~\citep{loshchilov2017decoupled}, with a learning rate of $1\mathrm{e}{-4} \text{, } \beta_1 = 0.9 \text{, } \beta_2 = 0.999 \text{, } \epsilon  = 1\mathrm{e}{-8}$.
\paragraph{Binary Classifier} This model is suited for the binary classification task of DAL and CAL. In this case it is a single $d_{\text{emb}}= $ 300 ReLu layer. We minimize the cross-entropy as well. 
We use the AdamW optimizer with a learning rate of $5\mathrm{e}{-5}$.
For more details regarding reproducibility, we refer to~\cref{reproducibility}. 

\paragraph{Training the Binary Classifier} 
For both the binary classification task of DAL and CAL,
we empirically determined that setting the number of epochs for the binary classifier to 30 yielded good results. In the context of DAL, for AGNews it reaches around 98\% accuracy during training, and for TREC it achieves around 89\% accuracy.
In contrast, with CAL, we start with little amounts of data for a binary classifier to train on. In the early stage of the AL iterations, the binary classifier does not achieve a high accuracy for both AGNews and TREC (around random). After it reaches the fifth or sixth AL iteration it starts to properly distinguish the \textbf{low-cor}/\textbf{hig-cor} samples, as it probably has enough samples to learn from. It achieves around 65--75\% accuracy for AGNews, and towards 85\% accuracy for TREC. The classification accuracy on AGNews seems low. However, further tuning of the binary classifier (e.g., increasing the number of epochs) slightly increases binary classification accuracy, but did not result in better performance for the overall AL setup.

\paragraph{Significance}
Recently, the Almost Stochastic Order test~\citep[ASO;\  ][]{dror2019deep}\footnote{Implementation of \citet{dror2019deep} can be found at~\url{https://github.com/Kaleidophon/deep-significance}~\cite{dennis_ulmer_2021_4638709}} has been proposed to test statistical significance for DNNs over multiple runs.
Generally, the ASO test determines whether a stochastic order~\citep{reimers2018comparing} exists between two models or algorithms based on their respective sets of evaluation scores. Given the single model scores over multiple random seeds of two algorithms $A$ and $B$, the method computes a test-specific value ($\epsilon_{min}$) that indicates how far algorithm $A$ is from being significantly better than algorithm $B$. When distance $\epsilon_{min} = 0.0$, one can claim that $A$ stochastically dominant over $B$ with a predefined significance level. When $\epsilon_{min} < 0.5$ one can say $A \succeq B$. On the contrary, when we have $\epsilon_{min} = 1.0$, this means $B \succeq A$. For $\epsilon_{min} = 0.5$, no order can be determined. We took 0.05 for the predefined significance level $\alpha$ and measure it across all 30 AL iterations.

\begin{figure*}[ht]
    \centering
    \includegraphics[width=.49\linewidth]{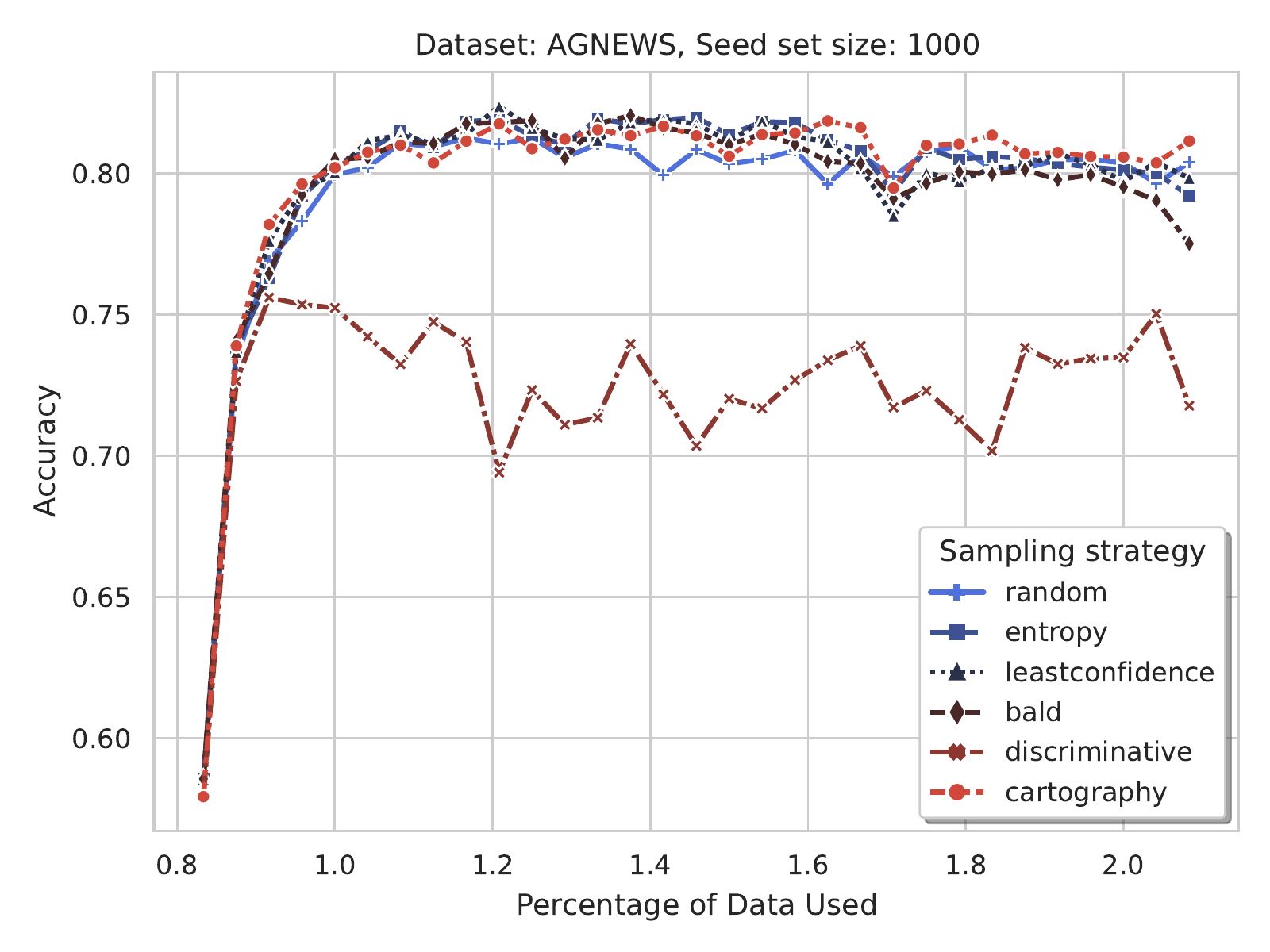}
    \includegraphics[width=.49\linewidth]{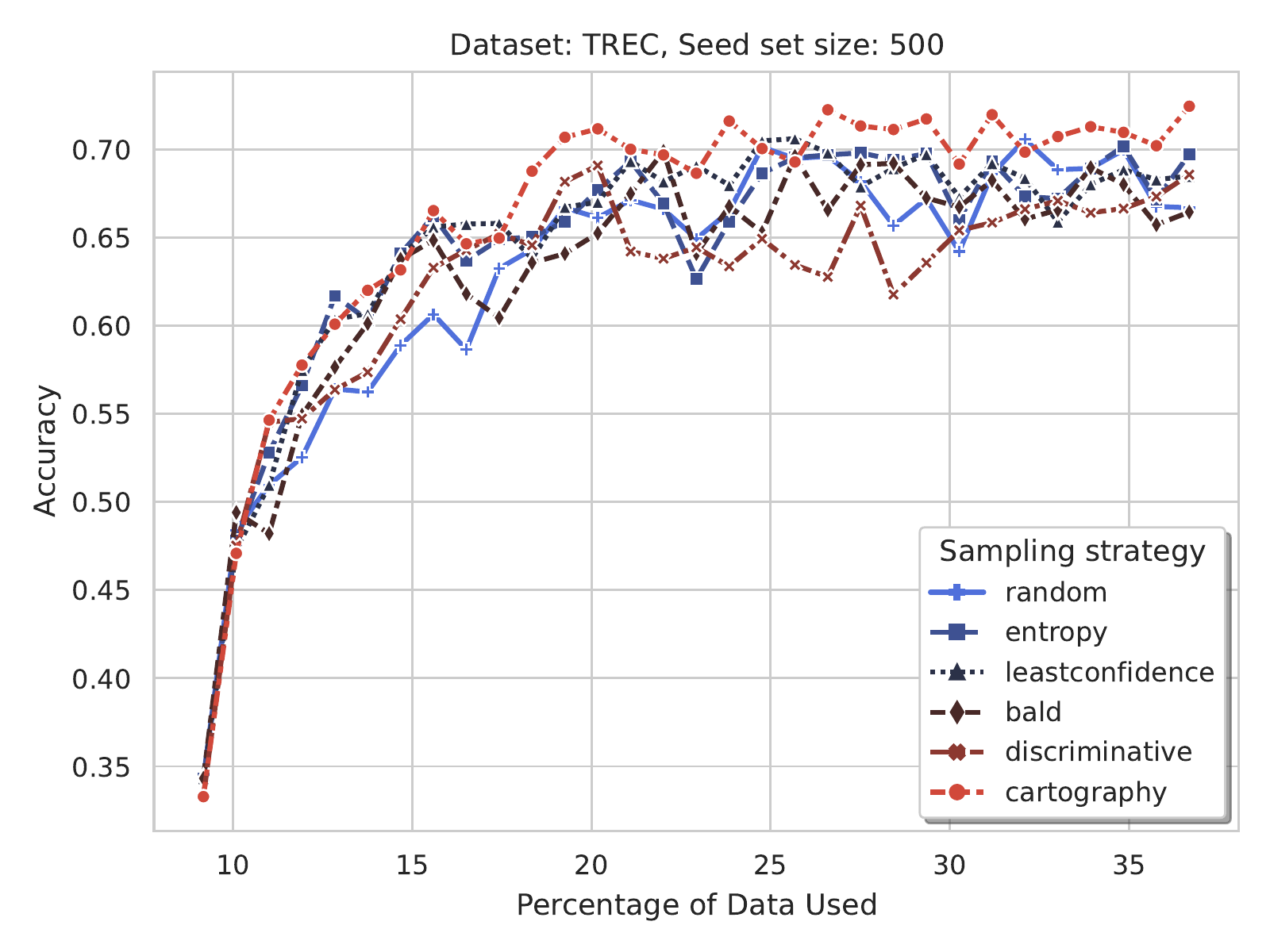}
    \caption{\textbf{Performance AL strategies.} Performance of the various AL strategies in terms of accuracy. The accuracy shown over the AL iterations is the average over five random seeds. Note that for both datasets we added the same number of instances to the seed set (+1,500 instances). The x-axis correspond to the fraction of the total size of the respective dataset. }
    \label{fig:trec-agnews-result}
\end{figure*}
\begin{table*}[ht]
    \centering
    \resizebox{.40\linewidth}{!}{
    \begin{tabular}{l|l|l|l|l|l|l}
    \toprule
    & \rotatebox[origin=c]{90}{Rand.} & \rotatebox[origin=c]{90}{LC} & \rotatebox[origin=c]{90}{Ent.} & \rotatebox[origin=c]{90}{BALD} & \rotatebox[origin=c]{90}{DAL} & \rotatebox[origin=c]{90}{CAL} \\
    \midrule
    Rand.     &  \cellcolor{gray}  &          1.00    & 1.00                & 1.00             & \textit{0.01}         & 1.00\\\hline
    LC        &  \textbf{0.00}     & \cellcolor{gray} & 1.00                & 1.00             & \textit{0.01}         & 1.00\\\hline
    Ent.      &  \textbf{0.00}     & \textbf{0.00}    & \cellcolor{gray}    & 1.00             & \textit{0.01}         & 1.00\\\hline
    BALD      &  \textbf{0.00}     & \textbf{0.00}    & 0.00                & \cellcolor{gray} & \textit{0.01}         & 1.00\\\hline
    DAL       &  0.99              & 0.99             & 0.99                & 0.99             & \cellcolor{gray}      & 1.00\\\hline
    CAL       &  \textbf{0.00}     & \textbf{0.00}    & \textbf{0.00}       & \textbf{0.00}    & \textbf{0.00}         & \cellcolor{gray}\\
    \bottomrule
    \end{tabular}}
    \hspace{2em}
    \resizebox{.40\linewidth}{!}{
    \begin{tabular}{l|l|l|l|l|l|l}
    \toprule
    & \rotatebox[origin=c]{90}{Rand.} & \rotatebox[origin=c]{90}{LC} & \rotatebox[origin=c]{90}{Ent.} & \rotatebox[origin=c]{90}{BALD} & \rotatebox[origin=c]{90}{DAL} & \rotatebox[origin=c]{90}{CAL} \\
    \midrule
    Rand.     &  \cellcolor{gray}& 1.00             & 1.00              & 1.00              & 1.00                  & 1.00\\\hline
    LC        &  \textbf{0.00}   & \cellcolor{gray} & 1.00              & 0.78              & 0.57                  & 1.00  \\\hline
    Ent.      &  \textbf{0.00}   & \textbf{0.00}    & \cellcolor{gray}  & 0.87              & 0.60                  & 1.00  \\\hline
    BALD      &  \textbf{0.00}   & \textit{0.22}    & \textit{0.13}     & \cellcolor{gray}  & 1.00                  & 1.00\\\hline
    DAL       &  \textbf{0.00}   & \textit{0.43}    & \textit{0.40}     & 1.00              & \cellcolor{gray}      & 1.00\\\hline
    CAL       &  \textbf{0.00}   & \textbf{0.00}    & \textbf{0.00}     & \textbf{0.00}     & \textbf{0.00}         & \cellcolor{gray}\\
    \bottomrule
    \end{tabular}}
    \caption{\textbf{Almost Stochastic Order Scores of AGNews (left) \& TREC (right).} ASO scores expressed in $\epsilon_{min}$ 
    The significance level $\alpha =$ 0.05 is adjusted accordingly by using the Bonferroni correction~\citep{bonferroni1936teoria}. \textbf{Bold} numbers indicate stochastic dominance and \textit{cursive} means that one algorithm is better than the other, e.g., for AGNews, LC (row) is stochastically dominant over the random baseline (column) with $\epsilon_{min}$ value of 0.00). Numbers are obtained across all 30 AL iterations}
    \label{tab:significance}
\end{table*}

\begin{figure*}[ht]
    \centering
    \includegraphics[width=\linewidth]{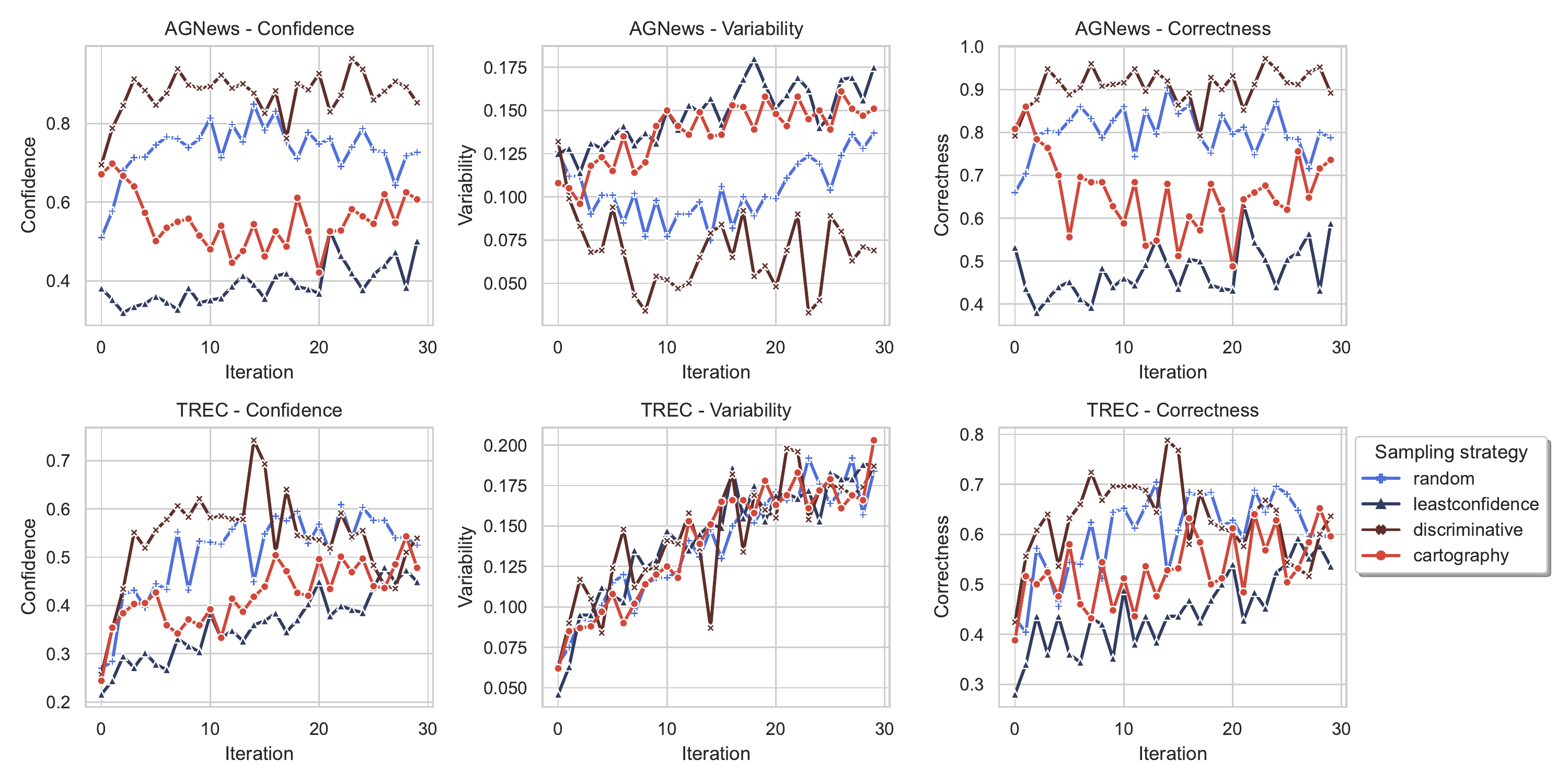}
    \caption{\textbf{Average Statistics AGNews \& TREC.}~Values correspond to the mean statistics~(\cref{cal}) of instances over the five random seeds. We calculate the statistics of each top-50 batch \textit{after} being added to the seed set. Therefore, we only have statistics of 29 runs, as in the 30th run we stop the AL cycle.}
    \label{fig:statistics-agnews-trec}
\end{figure*}

\section{Results \& Analysis}\label{sec:results}
We plot the accuracy of the AL algorithms~(\cref{subsec:al}) on each dataset in~\cref{fig:trec-agnews-result}. For AGNews and TREC, all AL strategies except DAL outperform the random baseline. CAL is statistically dominant over BALD (AGNews) and DAL (AGNews, TREC), and competitive with LC and Entropy (\cref{tab:significance}). This shows that CAL reaches strong results. CAL (illustrated as \texttt{cartography} in the figure with a red-dotted line).

\paragraph{Why does CAL work?}
To gain insight on why CAL works better, we investigate the average statistics of each selected batch of samples using the data maps. 
In~\cref{fig:statistics-agnews-trec}, we check the mean confidence, variability and correctness over each selected batch of 50 for sampling strategies Random, LC, DAL, and CAL for both AGNews and TREC. 
The statistics of the instances are extracted after the selected top--50 batch is added to the seed set. Once the main model is trained again on the increased seed set, we obtain the statistics of the previously added batch of 50.~\cref{fig:statistics-agnews-trec} shows that \texttt{leastconfidence} and \texttt{cartography}  has the higher variability amongst the other methods (middle graph). There is a similar signal as the findings of~\citet{swayamdipta-etal-2020-dataset}, the \textit{ambiguous} samples that the model \textit{learns the most} from are usually the instances that have the highest variability and average confidence---as we can see in~\cref{fig:statistics-agnews-trec} (especially pronounced on AGNews), these are the instances CAL selects.
In contrast, LC selects instances that have relatively low confidence and low variability in the early stages, but catches up in later ones. This suggests that LC chooses only \textit{hard-to-learn} instances at the start. In general, CAL seems to select more \textit{informative} samples as opposed to the random strategy.

DAL seems to start by choosing \textit{ambiguous} samples with high variability. However, it quickly picks mostly \textbf{high-cor} samples in later iterations, in contrast to CAL. Consequently, the performance for DAL drops as it leads to picking the \textit{easy-to-learn} samples. Picking too many \textit{easy-to-learn} instances results in worse optimization~\cite{swayamdipta-etal-2020-dataset}. The drop for DAL is visible in Figure~\ref{fig:trec-agnews-result}, which shows that CAL and DAL are close at start, and the accuracy of DAL then drops. 

\paragraph{What is the influence of the \textit{correctness} threshold?}

\begin{table}[htpb]
    \centering
    \resizebox{\linewidth}{!}{
    \begin{tabular}{l|rrrrr}
        \toprule
        $t_{cor}$      & $0.0$    & $\mathbf{\leq0.2}$ & $\leq0.4$ & $\leq0.6$ & $\leq0.8$\\
               \midrule
        AGNews & 0.789 & 0.811    & 0.796    & 0.788    & 0.781\\
        TREC   & 0.676 & 0.724    & 0.701    & 0.706    & --\\
        \bottomrule
    \end{tabular}}  
    \caption{\textbf{Influence of the Correctness Threshold.} 
    Influence of the correctness threshold $t_{cor}$ on the final average accuracy score per dataset over the five random seeds. $t_{cor}$ considers from what boundary we should consider \textbf{low-cor} cases. In \textbf{bold} is what threshold we are using.}
    \label{tab:threshold}
\end{table}

Here we investigate how changing the correctness threshold for the binary classification task impacts accuracy. As shown in~\cref{tab:threshold}, the selected threshold worked well for the two datasets studied. In particular, the accuracy does not drop substantially on AGNews if we move the \textit{correctness} threshold to a higher value.~\citet{swayamdipta-etal-2020-dataset} indicates that the \textit{ambiguous} region contains the instances with the highest variability. For AGNews, these instances have a \textit{correctness} range from 0.2--0.8 (as seen in~\cref{fig:trec-agnews}). Therefore, we assume that most of these instances are informative for the model.
In the case of TREC, the final accuracy also stays similar with different \textit{correctness} thresholds and does not drop substantially. We find in the full data map for TREC that the area with $t_{cor} = \{0.0, 0.4\}$ thresholds is more dense compared to $t_{cor} =$ 0.2 (details in~\cref{stats:full}). In other words, there are more samples, but also ones that are detrimental to performance. We note that the \textit{correctness} threshold could vary for different datasets and models.

\begin{table}
    \centering
        \resizebox{.7\linewidth}{!}{
\begin{tabular}{l|rr}
\toprule
                            & AGNews    & TREC\\
\midrule
CAL $\cap$ DAL              &  4                &  114 \\
CAL $\cap$ LC               &  2                &  101 \\
CAL $\cap$ Rand.            &  7                &  82 \\
\midrule
DAL $\cap$ LC               &  3                &  81 \\
DAL $\cap$ Rand.            &  2                &  87 \\
LC  $\cap$ Rand.            &  2                &  100\\
\bottomrule
\end{tabular}}
    \caption{\textbf{Number of Overlapping Instances on AGNews \& TREC.}~The values corresponds to the total overlapping instances out of 1,500 $*$ 5 random seeds $=$ 7,500.}
    \label{tab:overlap_agnews_trec}
\end{table}

\paragraph{Do AL strategies select the same instances for labeling?}\label{analysis:threshold}
We measured the overlap between the batches selected by each pair of strategies (Random, LC, DAL, CAL) on AGNews and TREC (\cref{tab:overlap_agnews_trec}). The batch overlap for CAL is low, with the highest overlap being 7 instances for AGNews with CAL $\cap$ Rand.\ followed by CAL $\cap$ DAL with 4 overlapping instances. The highest overlap for TREC is 114 instances for CAL $\cap$ DAL. Note this is the total overlap over five seeds. These results indicate that the AL algorithms choose different instances.

A popular approach for improving classification performance is combining (complementary) AL strategies. For example,~\citet{zhdanov2019diverse} proposed the idea to combine uncertainty sampling and diversity sampling for image classification.
As DAL and CAL have few overlapping instances, they can be complementary to each other. To test this, we combined DAL and CAL using a simple heuristic, by providing both of them half of the annotation budget (i.e., take top-25 batch of each AL strategy). This resulted in an accuracy score of 0.695 for TREC and 0.789 for AGNews. This suggests that it could have a positive effect if there is a more sophisticated approach. This is an open research topic that requires further investigation.

\paragraph{How data-efficient is CAL in comparison to full data training?} If a model is able to choose the instances that it can learn the most from, it can reach comparable results or even outperform a model trained on all data by using fewer training instances. This is noted by~\citet{siddhant2018deep}, where they achieve 98--99\% of the full dataset performance while labeling only 20\% of the samples. The overall accuracy for AGNews trained on all data is 0.820 accuracy on test. For TREC, this results in 0.634. With CAL, we achieve around 99\% of the full dataset performance while using only 2\% training data for AGNews. For TREC, we outperform the full dataset performance by 0.090 accuracy (0.634 vs.\ 0.724), the full dataset performance is already reached by using around 14\% training data. This is appealing, as active learning can provide more data-effective learning solutions.

\section{Conclusion}
In this paper, we introduced a new AL algorithm, Cartography Active Learning. The AL objective is transformed into a binary classification task~\citep{gissin2019discriminative}, where we optimize for selecting the most informative data with respect to a model by leveraging insights from data maps~\citep{swayamdipta-etal-2020-dataset}.  Data maps help to identify distinct regions in a dataset based on training dynamics (\textit{hard-to-learn} and \textit{easy-to-learn/ambiguous} instances), which have shown to play an important role in model optimization and stability in full dataset training~\cite{swayamdipta-etal-2020-dataset}. We use these insights in low-data regimes and propose CAL. 
Specifically for the task of Visual Question Answering, contemporary work by~\citet{karamcheti-etal-2021-mind} use data maps for analysis into AL acquisition functions and the effect of outliers. In contrast, this work instead applies data maps directly for AL by training a discriminator on limited seed data maps to select the most informative instances.
We show empirically that our method is competitive or significantly outperforms various popular AL methods, and provide intuitions on why this is the case by using training dynamics.  

\section*{Acknowledgements}
\noindent We thank the NLPnorth group for feedback on an earlier version of this paper --- in particular Elisa Bassignana and Max M{\"u}ller-Eberstein for insightful discussions. We would also like to thank the anonymous reviewers for their comments to improve this paper. Last, we also thank NVIDIA and the ITU High-performance Computing cluster for computing resources.  This research is supported by the Independent Research Fund Denmark (DFF) grant 9131-00019B. 

\bibliography{anthology,output}
\bibliographystyle{acl_natbib}

\clearpage
\section*{Appendix}

\section{Reproducibility}\label{reproducibility}
We initialize the main model with English $d_{\text{emb}}= $ 300 FastText embeddings~\citep{bojanowski2017enriching} and keep it frozen during training and inference, we sum the embeddings over the sequence of tokens as motivated by~\citet{banea-etal-2014-simcompass}. For AGNews we impose a maximum sequence length of 200 and a batch size of 64. For TREC, a maximum sentence length of 42 and a batch size of 16. We run both models for 30 epochs, with no early stopping. Per AL iteration, we do a weight reset on all models. We average our results over five randomly generated seeds (398048, 127003, 259479, 869323, 570852). All experiments were ran on an NVIDIA\textsuperscript{\textregistered} A100 SXM4 40 GB GPU
and an AMD EPYC\textsuperscript{\texttrademark} 7662 64-Core CPU. Specifically for CAL, a single AL batch (50) selection iteration takes 11 seconds on average assuming TREC. For AGNews, one AL iteration takes 62 seconds on average. Both runtimes are with respect to the models depicted in~\cref{configurations} and hardware mentioned above.

\section{Full Data Map}\label{stats:full}

\begin{figure*}[!htpb]
    \centering
    \includegraphics[width=0.9\linewidth]{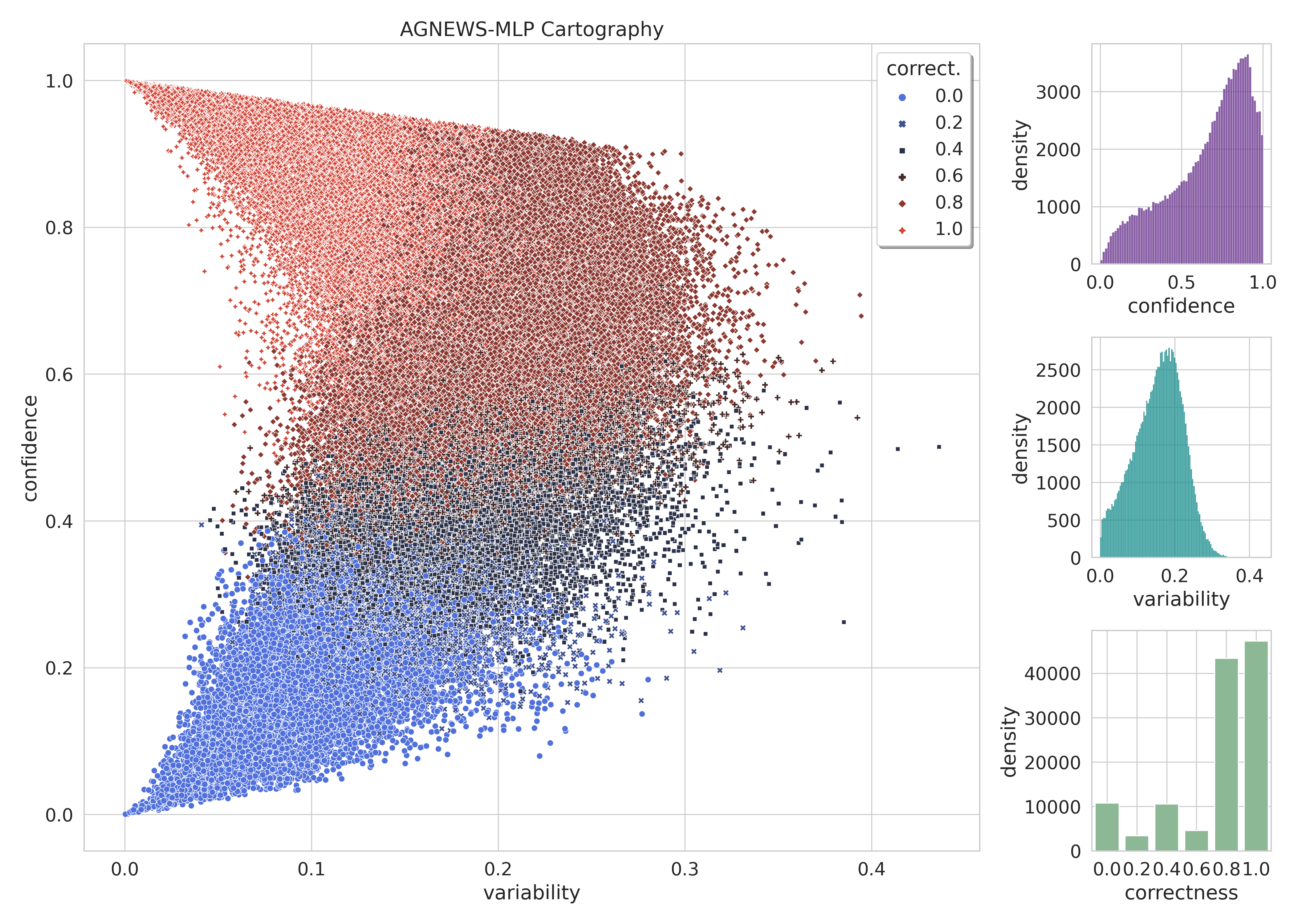}
    \caption{\textbf{Density of AGNews.} Density statistics of AGNews over ten epochs.}
    \label{fig:agnews-results}
\end{figure*}

\begin{figure*}[!htpb]
    \centering
    \includegraphics[width=0.9\linewidth]{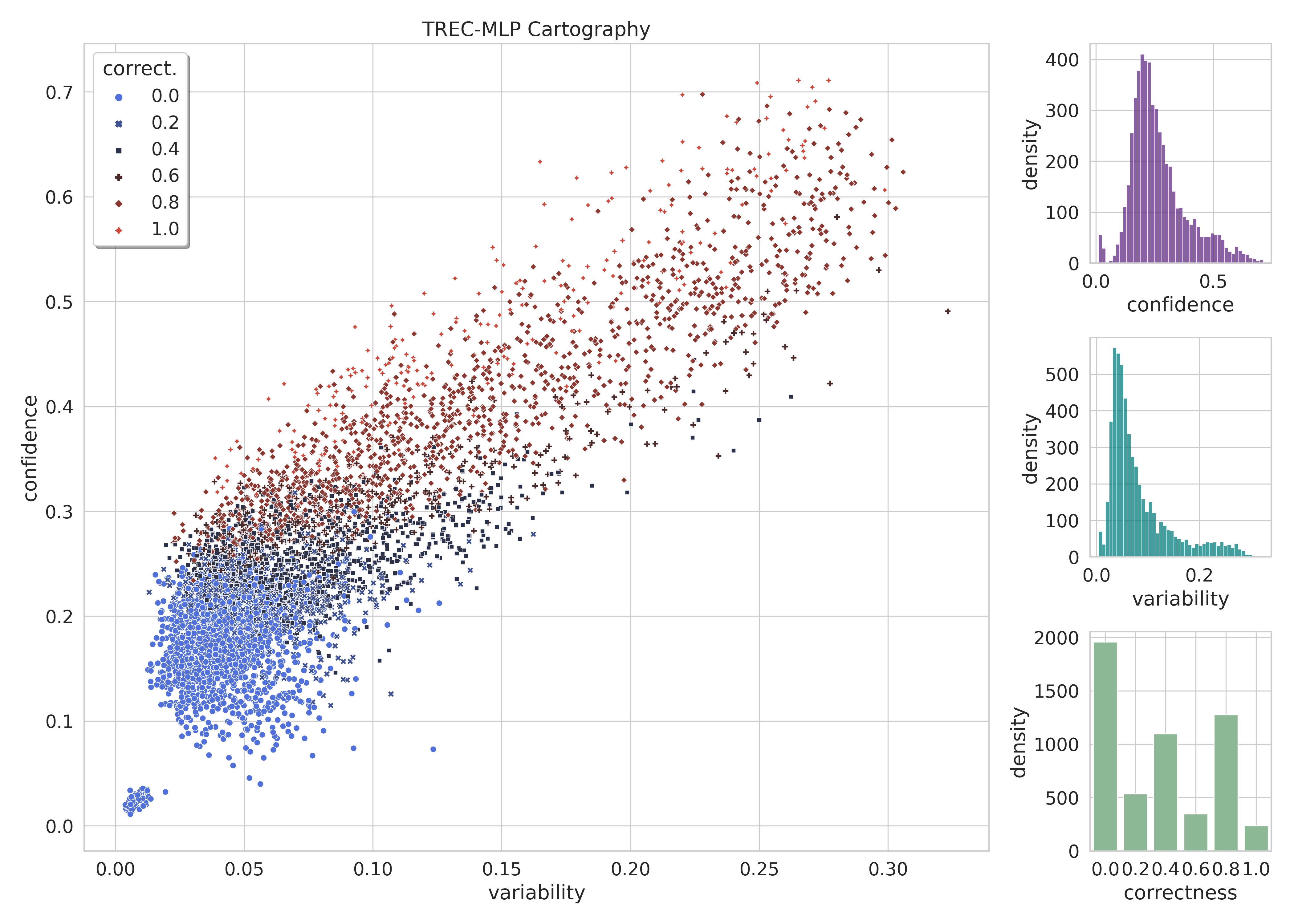}
    \caption{\textbf{Density of TREC.} Density statistics of TREC over ten epochs.}
    \label{fig:trec-results}
\end{figure*}

\cref{fig:agnews-results} and~\cref{fig:trec-results} show the full data maps for AGNews and TREC respectively. Identically to~\citet{swayamdipta-etal-2020-dataset}, we show the density of data points in the plots. We can see a clear difference in density between the datasets. For AGNews, we can see the majority of data points have a high confidence ($\sim$0.8) and high correctness. In contrast, TREC contains plenty of instances that have low confidence ($\sim$0.3) and low correctness. 

\end{document}